\documentclass{article}

\usepackage{microtype}
\usepackage{graphicx}
\usepackage{subfigure}
\usepackage{booktabs}
\usepackage{hyperref}

\newcommand{\X}{\mathcal{X}}
\newcommand{\SSS}{\mathcal{S}}
\newcommand{\AAA}{\mathcal{A}}
\newcommand{\XXX}{\X}
\newcommand{\TTT}{\mathcal{T}}
\newcommand{\LL}{\mathcal{L}}
\newcommand{\RR}{\mathbb{R}}
\newcommand{\ZZ}{\mathbb{Z}}
\newcommand{\ra}{\rightarrow}

\usepackage[accepted]{icml2023}

\usepackage{amsmath}
\usepackage{amssymb}
\usepackage{mathtools}
\usepackage{amsthm}

\usepackage[capitalize,noabbrev]{cleveref}

\theoremstyle{plain}

\theoremstyle{definition}

\theoremstyle{remark}

\makeatletter
\renewcommand{\paragraph}[1]{\textbf{#1}\ \ }
\makeatother

\icmltitlerunning{Thompson Sampling for Improved Exploration in GFlowNets}

\begin{document}

\twocolumn[
\icmltitle{Thompson Sampling for Improved Exploration in GFlowNets}

\icmlsetsymbol{equal}{*}

\begin{icmlauthorlist}
\icmlauthor{Jarrid Rector-Brooks}{mila,udem,dreamfold}
\icmlauthor{Kanika Madan}{mila,udem}
\icmlauthor{Moksh Jain}{mila,udem}
\icmlauthor{Maksym Korablyov}{mila,dreamfold}
\icmlauthor{Cheng-Hao Liu}{mila,dreamfold,mcgill}
\icmlauthor{Sarath Chandar}{mila,poly}
\icmlauthor{Nikolay Malkin}{mila,udem}
\icmlauthor{Yoshua Bengio}{mila,udem,cifar}
\end{icmlauthorlist}

\icmlaffiliation{mila}{Mila -- Qu\'ebec AI Institute}
\icmlaffiliation{udem}{Universit\'e de Montr\'eal}
\icmlaffiliation{poly}{Polytechnique Montr\'eal}
\icmlaffiliation{mcgill}{McGill University}
\icmlaffiliation{cifar}{CIFAR Fellow}
\icmlaffiliation{dreamfold}{DreamFold}

\icmlcorrespondingauthor{Jarrid Rector-Brooks}{jarrid.rector-brooks@mila.quebec}

\icmlkeywords{Machine Learning, ICML}

\vskip 0.3in
]

\printAffiliationsAndNotice{} 

\begin{abstract}
Generative flow networks (GFlowNets) are amortized variational inference algorithms that treat sampling from a distribution over compositional objects as a sequential decision-making problem with a learnable action policy. Unlike other algorithms for hierarchical sampling that optimize a variational bound, GFlowNet algorithms can stably run off-policy, which can be advantageous for discovering modes of the target distribution. Despite this flexibility in the choice of behaviour policy, the optimal way of efficiently selecting trajectories for training has not yet been systematically explored. In this paper, we view the choice of trajectories for training as an active learning problem and approach it using Bayesian techniques inspired by methods for multi-armed bandits. The proposed algorithm, Thompson sampling GFlowNets (TS-GFN), maintains an approximate posterior distribution over policies and samples trajectories from this posterior for training. We show in two domains that TS-GFN yields improved exploration and thus faster convergence to the target distribution than the off-policy exploration strategies used in past work.
\end{abstract}

\section{Introduction}
\label{intro}
Generative flow networks \citep[GFlowNets;][]{bengio2021flow} are generative models which sequentially construct objects from a space $\X$ by taking a series of actions sampled from a learned policy $P_F$.  A GFlowNet's policy $P_F$ is trained such that, at convergence, the probability of obtaining some object $x \in \X$ as the result of sampling a sequence of actions from $P_F$ is proportional to a reward $R(x)$ associated to $x$.  Whereas traditional probabilistic modeling approaches (e.g., those based on Markov chain Monte Carlo (MCMC)) rely on local exploration in $\X$ for good performance, the parametric policy learned by GFlowNets allows them to generalize across states and yield superior performance on a number of tasks \citep{bengio2021flow,malkin2022trajectory,zhang2022generative,jain2022biological,deleu2022bayesian,jain2022multiobjective,gfn-em,zhang2023robust}.

While GFlowNets solve the variational inference problem of approximating a target distribution on $\X$ with the distribution induced by the sampling policy \citep{malkin2022gfnhvi}, they are trained in a manner reminiscent of reinforcement learning (RL).  GFlowNets are typically trained by either sampling trajectories on-policy from the learned sampling policy or off-policy from a mix of the learned policy and random noise.  Each trajectory sampled concludes with some object $x \in \X$ for which the GFlowNet receives reward $R(x)$ and takes a gradient step on the parameters of the sampler with respect to the reward signal.  Despite GFlowNets' prior successes, this mode of training leaves them vulnerable to issues seen in the training of reinforcement learning agents --- namely, slow temporal credit assignment and optimally striking the balance between exploration and exploitation.

Although multiple works have tackled the credit assignment issue in GFlowNets \citep{malkin2022trajectory,madan2022learning,deleu2022bayesian,pan2023better}, considerably less attention has been paid to the exploration problem.  Recently \citet{pan2022gafn} proposed to augment GFlowNets with intermediate rewards so as to allow the addition of intrinsic rewards \citep{pathak2017curiosity,burda2018exploration} and incorporate an exploration signal directly into training.  However, while density-based exploration bonuses can provide much better performance on tasks where the reward $R(x)$ is very sparse, there is no guarantee that the density-based incentives correlate with model uncertainty or task structure. In fact, they have been shown to yield arbitrarily poor performance in a number of reinforcement learning settings \citep{osband2019deep}. In this paper, we develop an exploration method for GFlowNets which provides improved convergence to the target distribution \textit{even when the reward $R(x)$ is not sparse.}

Thompson sampling \citep[TS;][]{thompson1933likelihood} is a method which provably manages the exploration/exploitation problem in settings from multi-armed bandits to reinforcement learning \citep{agrawal2017optimistic,agrawal2017near} and has been employed to much success across a variety of deep reinforcement learning tasks \citep{osband2016deep,osband2018randomized,osband2019deep}.  The classical TS algorithm~\citep{agrawal2012analysis,russo2018tutorial} maintains a posterior over the model of the environment and acts optimally according to a sample from this posterior over models. TS has been generalized to RL problems in the form of Posterior Sampling RL~\citep{osband2013more}. A variant of TS has been adapted in RL, where the agent maintains a posterior over policies and value functions~\citep{osband2016generalization,osband2016deep} and acts optimally based on a random sample from this posterior. We consider this variant of TS in this paper. 

\textbf{Our main contribution in this paper is describing and evaluating an algorithm based on Thompson sampling for improved exploration in GFlowNets}.  Building upon prior results in \citet{malkin2022trajectory,madan2022learning} we demonstrate how Thompson sampling with GFlowNets allows for improved exploration and optimization efficiency in GFlowNets.  We validate our method on a grid-world and sequence generation task.  In our experiments TS-GFN substantially improves both the sample efficiency and the task performance.  Our algorithm is computationally efficient and highly parallelizable, only taking $\sim15\%$ more computation time than prior approaches.

\section{Related Work}
\label{related_work}
\paragraph{Exploration in RL}
There exists a wide literature on uncertainty based RL exploration methods.  Some methods rely on the Thompson sampling heuristic and non-parametric representations of the posterior to promote exploration \citep{osband2013more,osband2016deep,osband2016generalization,osband2018randomized}.  Others employ uncertainty to enable exploration based on the upper confidence bound heuristic or information gain \citep{ciosek2019better,lee2021sunrise,o2018uncertainty,nikolov2018information}.
Another set of exploration methods attempts to make agents ``intrinsically'' motivated to explore.  This family of methods includesrandom network distillation (RND) and Never Give Up \citep{burda2018exploration,badia2020never}.  \citet{pan2022gafn}, proposes to augment GFlowNets with intrinsic  RND-based intrinsic rewards to encourage better exploration.

\paragraph{MaxEnt RL} RL has a rich literature on energy-based, or maximum entropy, methods~\citep{ziebart2010modeling,mnih2016asynchronous,haarnoja2017reinforcement,nachum2017bridging,schulman2017proximal,haarnoja2018soft}, which are close or equivalent to the GFlowNet framework in certain settings (in particular when the MDP has a tree structure~\citep{bengio2021flow}). Also related are methods that maximize entropy of the state visitation distribution or some proxy of it~\citep{hazan2019provably,islam2019marginalized,zhang2021exploration,eysenbach2018diversity}, which achieve a similar objective to GFlowNets by flattening the state visitation distribution.  We hypothesize that even basic exploration methods for GFlowNets (e.g., tempering or $\epsilon$-noisy) could be sufficient exploration strategies on some tasks.

\section{Method}
\label{method}
\subsection{Preliminaries}
\label{subsec:prelims}
We begin by summarizing the preliminaries on GFlowNets, following the conventions of \citet{malkin2022trajectory}.

Let $G=(\SSS,\AAA)$ be a directed acyclic graph. The vertices $s\in\SSS$ are called \emph{states} and the directed edges $(u\ra v)\in\mathcal{A}$ are \emph{actions}. If $(u\ra v)$ is an edge, we say $v$ is a \emph{child} of $u$ and $u$ is a \emph{parent} of $v$. There is a unique \emph{initial state} $s_0\in\SSS$ with no parents. States with no children are called \emph{terminal}, and the set of terminal states is denoted by $\XXX$.

A \emph{trajectory} is a sequence of states $\tau=(s_m\ra s_{m+1}\ra\dots\ra s_n)$, where each $(s_i\ra s_{i+1})$ is an action. The trajectory is \emph{complete} if $s_m=s_0$ and $s_n$ is terminal. 
The set of complete trajectories is denoted by $\TTT$.

A \emph{(forward) policy} is a collection of distributions $P_F(-|s)$ over the children of every nonterminal state $s\in\SSS$. A forward policy determines a distribution over $\TTT$ by
\begin{equation}
    P_F(\tau=(s_0\ra\dots\ra s_n))=\prod_{i=0}^{n-1}P_F(s_{i+1}|s_i).
\end{equation}
Similarly, a \emph{backward policy} is a collection of distributions $P_B(-|s)$ over the \emph{parents} of every noninitial state.

Any distribution over complete trajectories that arises from a forward policy satisfies a Markov property: the marginal choice of action out of a state $s$ is independent of how $s$ was reached. Conversely, any Markovian distribution over $\TTT$ arises from a forward policy~\citep{bengio2021foundations}.

A forward policy can thus be used to sample terminal states $x\in\XXX$ by starting at $s_0$ and iteratively sampling actions from $P_F$, or, equivalently, taking the terminating state of a complete trajectory $\tau\sim P_F(\tau)$. The marginal likelihood of sampling $x\in\XXX$ is the sum of likelihoods of all complete trajectories that terminate at $x$.

Suppose that a nontrivial (not identically 0) nonnegative reward function $R:\XXX\to\RR_{\geq0}$ is given. The learning problem solved by GFlowNets is to estimate a policy $P_F$ such that the likelihood of sampling $x\in\XXX$ is proportional to $R(x)$. That is, there should exist a constant $Z$ such that
\begin{equation}
    R(x)=Z\sum_{\tau in \TTT: \tau=(s_0\ra\dots\ra s_n=x)}P_F(\tau)\quad\forall x\in\XXX.
    \label{eqn:reward_sampling}
\end{equation}
If (\ref{eqn:reward_sampling}) is satisfied, then $Z=\sum_{x\in\XXX}R(x)$. 
The sum in (\ref{eqn:reward_sampling}) may be intractable. Therefore, GFlowNet training algorithms require estimation of auxiliary quantities beyond the parameters of the policy $P_F$. The training objective we primarily consider, \emph{trajectory balance} (TB), learns an estimate of the constant $Z$ and of a \emph{backward policy}, $P_B(s\mid s')$, representing the posterior over predecessor states of $s'$ in trajectories that contain $s'$. The TB loss for a trajectory $\tau$ is:
\begin{equation}
    \LL_{TB}(\tau; \theta) = \left(\log \frac{Z_\theta \prod_{t=0}^{n-1} P_F(s_{t+1}|s_t; \theta)}{R(s_n) \prod_{t=0}^{n-1} P_B(s_t|s_{t+1}; \theta)}\right)^2
    \label{eq:tb_loss}
\end{equation}
where $\theta$ are the parameters of the learned objects $P_F$, $P_B$, and $Z$. If $\LL_{TB}(\tau; \theta)=0$ for all $\tau$, then $P_F$ samples objects $x\in\XXX$ with probability proportional to $R(x)$, i.e., (\ref{eqn:reward_sampling}) is satisfied. Algorithms minimize this loss for trajectories $\tau$ sampled from some \emph{training policy} $\pi_\theta$, which may be equal to $P_F$ itself (\emph{on-policy training}) but is usually taken to be a more exploratory distribution, as we discuss below.

 Notably, any choice of a backwards policy $P_B$ yields a unique corresponding $P_F$ and $Z$ which makes the expression on the right side of (\ref{eq:tb_loss}) equal to zero for all $\tau \in \TTT$ (see \citet{malkin2022gfnhvi} for interpretations of this result in terms of variational methods). 

\subsection{GFlowNet exploration strategies}
Prior work on GFlowNets uses training policies based on dithering or intrinsic motivation, including:

\begin{description}
    \item[On-policy] The training policy is the current $P_F$: $\pi_\theta(s' | s) = P_F(s' | s;\theta)$.
    \item[Tempering] Let $\alpha_\theta(s'|s): \SSS \times \SSS \ra \RR$ be the logits of $P_F$, then the training policy is a Boltzmann distribution with temperature $T \in \RR$ as
    $\pi_\theta(s'|s) \propto \exp\left(\alpha_\theta(s'|s) / T\right)$.
    \item[$\epsilon$-noisy] For $\epsilon \in [0,1]$, the training policy follows $P_F$ with probability $1 - \epsilon$ and takes a random action with probability $\epsilon$ as $\pi_\theta(s'|s) = (1-\epsilon)P_F(s'|s;\theta) + \frac{\epsilon}{\#\{s'':(s\ra s'')\in\AAA\}}$.
    \item[GAFN \citep{pan2022gafn}] The training policy is the current $P_F$, but $P_F$ is learned by incorporating a pseudocount-based intrinsic reward for each state $s \in \tau$ into the objective $\LL(\tau;P_F,P_B)$ so that
    $\pi_\theta(s'|s) = P_F(s'|s;\theta)$.
\end{description}

\subsection{Thompson sampling for GFlowNets}
Learning GFlowNets over large spaces $\XXX$ requires judicious exploration.  It makes little sense to explore in regions the GFlowNet has already learned well -- we would much rather prioritize exploring regions of the state space on which the GFlowNet has not accurately learned the reward distribution.  Prior methods do not explicitly prioritize this.  Both dithering approaches (tempering and $\epsilon$-noisy) 
and GAFNs encourage a form of uniform exploration, be it pure random noise as in dithering or a pseudocount in GAFNs.  While it is impossible to \emph{a priori} determine which regions a GFlowNet has learned poorly, we might expect that it performs poorly in the regions on which it is uncertain. An agent with an estimate of its own uncertainty could bias its action selection towards regions in which it is more uncertain.

With this intuition in mind, we develop an algorithm inspired by Thompson sampling and its applications in RL and bandits \citep{osband2016deep,osband2018randomized}.  In particular, following \citet{osband2016deep} we maintain an approximate posterior over forward policies $P_F$ by viewing the last layer of our policy network itself as an ensemble.  To maintain a size $K \in \ZZ^+$ ensemble extend the last layer of the policy network to have $K \cdot \ell$ heads where $\ell$ is the maximum number of valid actions according to $G$ for any state $s \in \SSS$.  To promote computational efficiency all members of our ensemble share weights in all layers prior to the final one.  

To better our method's uncertainty estimates, we employ the statistical bootstrap to determine which trajectories $\tau$ may be used to train ensemble member $P_{F,k}$ and also make use of randomized prior networks \citep{osband2018randomized}.  Prior networks are a downsized version of our main policy network whose weights are fixed at initialization and whose output is summed with the main network in order to produce the actual policy logits.  Prior networks have been shown to significantly improve uncertainty estimates and agent performance in reinforcement learning tasks.  

Crucially, while we parameterize an ensemble of $K$ forward policies we do not maintain an ensemble of backwards policies, instead sharing one $P_B$ across all ensemble members $P_{F,k}$.  Recall from \ref{subsec:prelims} that each $P_B$ uniquely determines a $P_F$ which $\mathcal{L}_{TB}(\tau) = 0 \quad \forall \tau \in \TTT$.  Specifying a different $P_{B,k}$ for each $P_{F,k}$ would result in setting a different learning target for each $P_{F,k}$ in the ensemble.  By sharing a single $P_B$ across all ensemble members we ensure that all $P_{F,k}$ converge to the same optimal $P_F^*$.  We show in Section \ref{exp:grid} that sharing $P_B$ indeed yields significantly better performance than maintaining separate $P_{B,k}$.

With our policy network parameterization in hand, the rest of our algorithm is simple. First we sample an ensemble member $P_{F,k}$ with $k \sim \textrm{Uniform}\{1,\dots,K\}$ and then sample an entire trajectory from it $\tau \sim P_{F,k}$.  This trajectory is then used to train each ensemble member where we include the trajectory in the training batch for ensemble member $P_{F,k}$ based on the statistical bootstrap with bootstrap probability $p$ ($p$ is a hyperparameter fixed at the beginning of training).  The full algorithm is presented in Appendix \ref{app:algo}.
\section{Experiments}
\label{experiments}
\subsection{Grid}
\label{exp:grid}
\begin{figure}[t]
    \centering
    \includegraphics[width=0.26\textwidth,trim=0 40 0 40,clip]{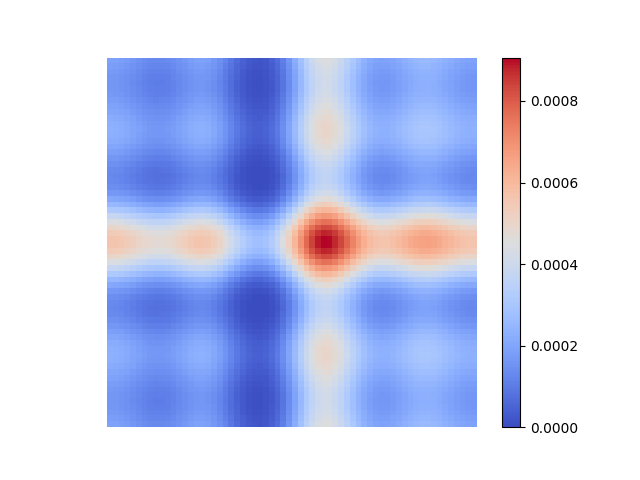}
    \includegraphics[width=0.16\textwidth,trim=0 0 0 10,clip]{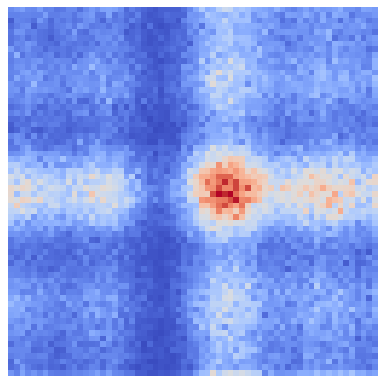}
    \caption{Reward on the grid task. \textbf{Left:} true distribution (normalized reward function). \textbf{Right:} empirical distribution over last $2 \cdot 10^5$ states sampled from the GFlowNet at end of training.}
    \label{fig:grid_reward_shape}
\end{figure}
We study a modified version of the grid environment from \citep{bengio2021flow}.  The set of interior states is a 2-dimensional grid of size $H \times H$.  The initial state is $(0, 0)$ and each action is a step that increments one of the 2 coordinates by 1 without leaving the grid.  A special termination action is also allowed from each state.

Prior versions of this grid environment provide high reward whenever the agent exits at a corner of the grid.  This sort of reward structure is very easy for an agent to generalize to and is a trivial exploration task when the reward is not highly sparse (such reward structures are \emph{not} the focus of this paper).  To compensate for this, we adopt a reward function based on a summation of truncated Fourier series, yielding a reward structure which is highly multimodal and more difficult to generalize to (see Figure \ref{fig:grid_reward_shape}).  The reward function is given by
\begin{align*}
    R(x) = \sum_{k=1}^n &\cos(2a_{k,1}\pi g(x_1)) + \sin(2 a_{k,2} \pi g(x_1)) + \\
    &\cos(2b_{k,1}\pi g(x_2)) + \sin(2b_{k,2}\pi g(x_2)) 
\end{align*}
where $a_{k,1},a_{k,2},b_{k,1},b_{k,2} \in \RR$ are preset scaling constants $\forall k$, $n$ is a hyperparameter determining the number of elements in the summation, $g: \ZZ_{\geq 0} \ra [c,d], g(x) = \frac{x(d-c)}{H} + c$, and $c, d \in \RR$ are the first and last integer coordinates in the grid.

We investigate a $64 \times 64$ grid with this truncated Fourier series reward (see Appendix \ref{app:grid} for full reward setup details).  We train the GFlowNets to sample from this target reward function and plot the evolution of the $L_1$ distance between the target distribution and the empirical distribution of the last $2 \cdot 10^5$ states seen in training\footnote{This evaluation is possible in this environment because the exact target distribution can be tractably computed.}.

\begin{figure}[t]
    \centering
    \includegraphics[width=0.235\textwidth]{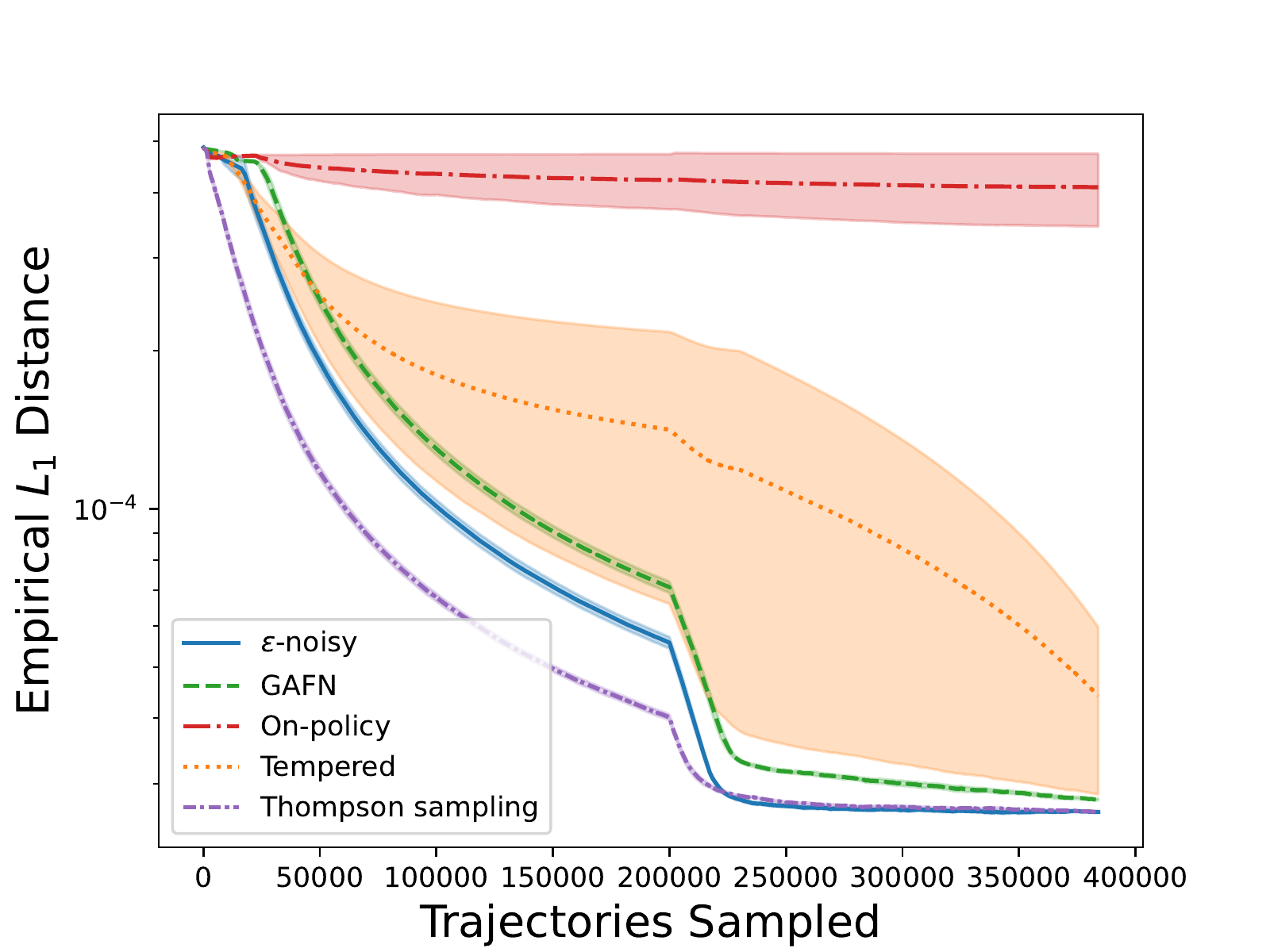}
    \includegraphics[width=0.235\textwidth]{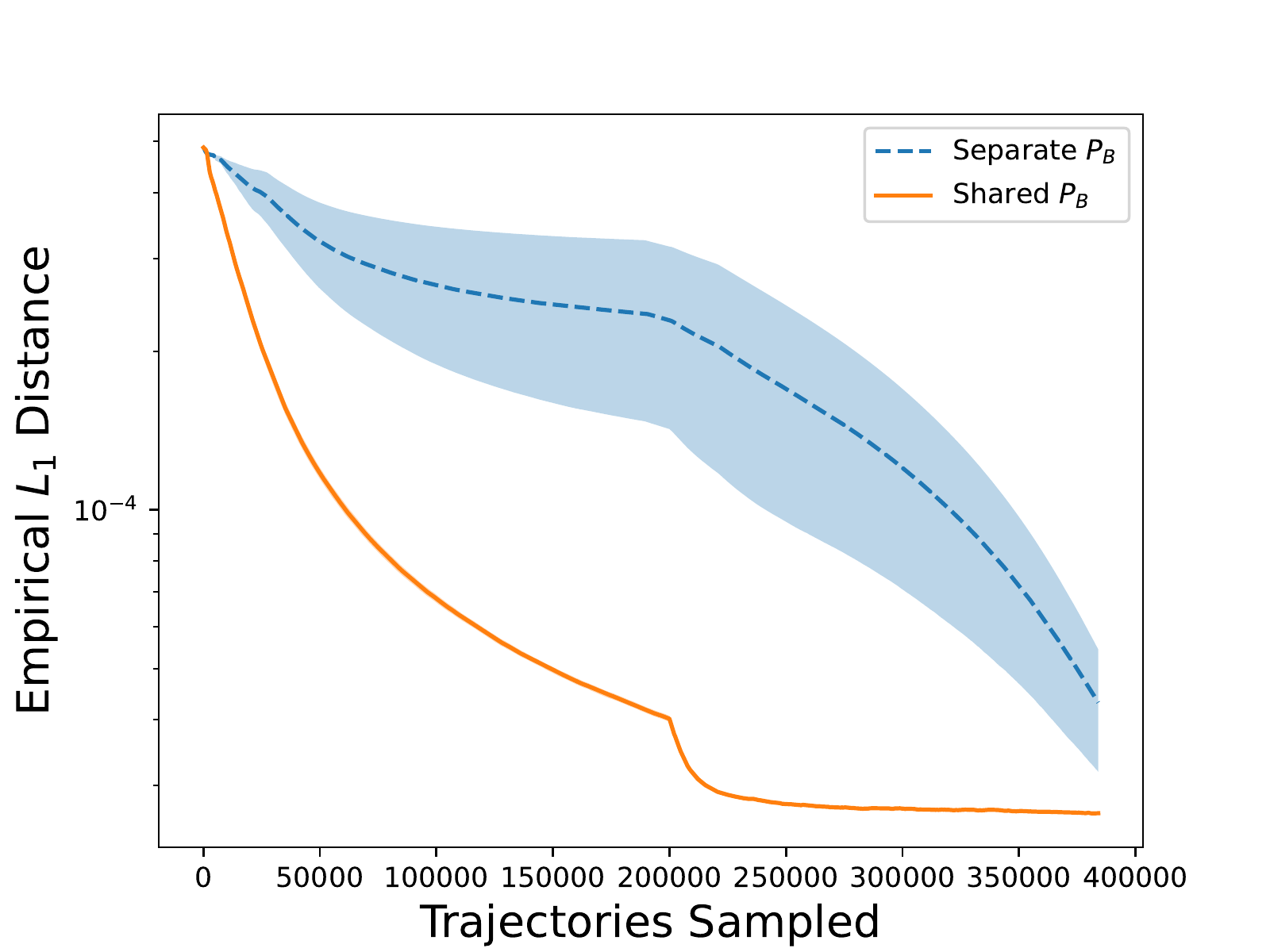}
    \caption{$L_1$ distance between empirical and target distributions over the course of training on the hypergrid environment (mean is plotted with standard error bars over 5 random seeds). \textbf{Left:} Thompson sampling learns the distribution better and faster than all other methods. \textbf{Right:} sharing a backwards policy $P_B$ performs significantly better than maintaining a separate backward policy $P_{B,k}$ for each forward policy $P_{F,k}$ in the ensemble.}
    \label{fig:grid_performance}
\end{figure}

The results (mean and standard error over five random seeds) are shown in Figure \ref{fig:grid_performance} (left side).  Models trained with trajectories sampled by TS-GFN converge faster and with very little variance over random seeds to the true distribution than all other exploration strategies.

We also investigate the effect of sharing the backwards policy $P_B$ across ensemble members in Figure \ref{fig:grid_performance} (right side).  Maintaining a separate $P_{B,k}$ for each $P_{F,k}$ performs significantly worse than sharing a single $P_B$ over all ensemble members.  Maintaining separate $P_{B,k}$ resulted in the GFlowNet learning much slower than sharing $P_B$ and converging to a worse empirical $L_1$ than sharing $P_B$.

\subsection{Bit sequences}
\begin{figure}[t]
    \centering
    \includegraphics[width=0.4\textwidth]{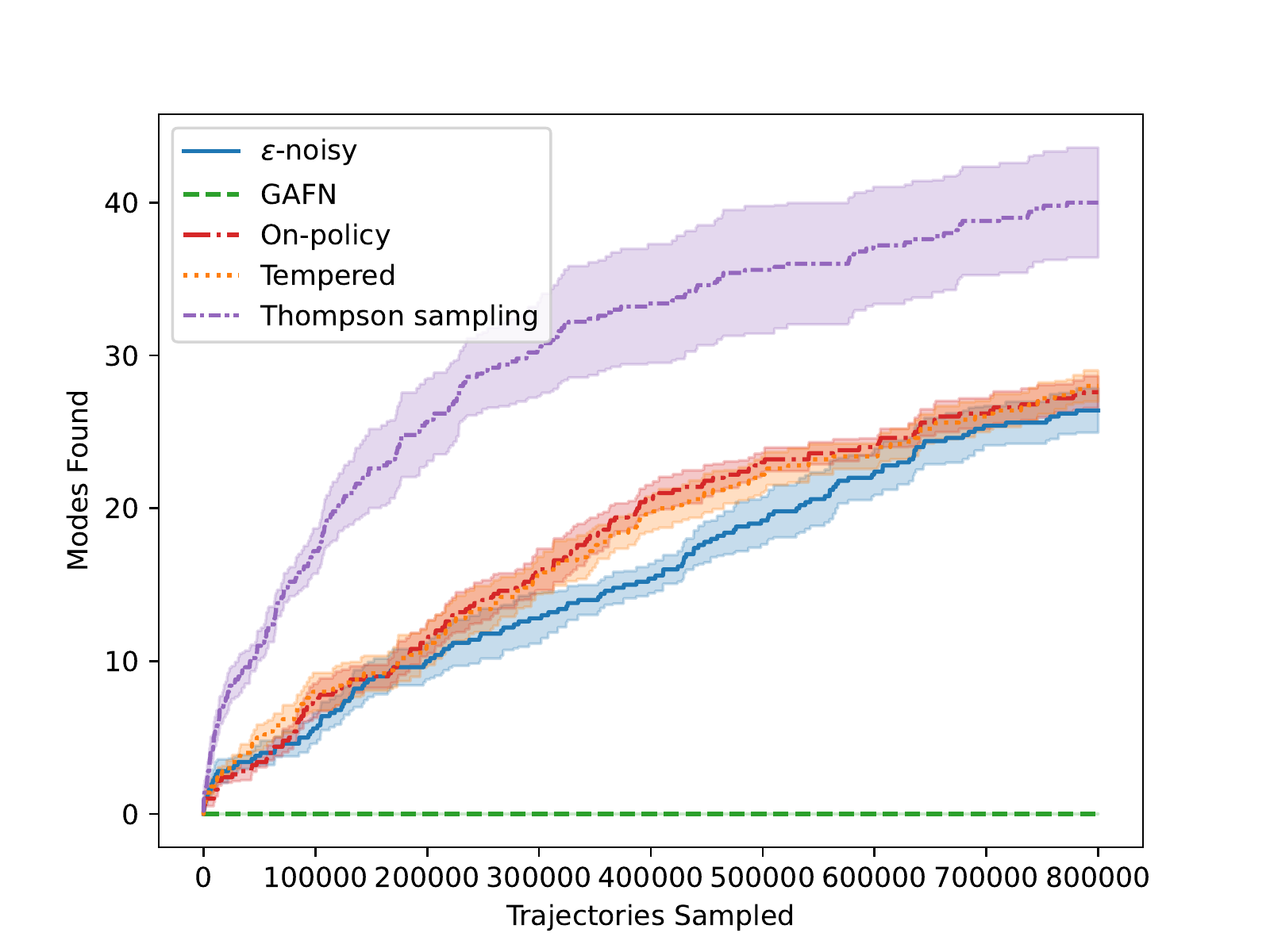}\vspace{-1em}
    \caption{Number of modes found as a function of training time for bit sequence task.}
    \label{fig:bit_seq_1_modes_found}
\end{figure}
We consider the synthetic sequence generation setting from \citet{malkin2022trajectory}, where the goal is to generate sequences of bits of fixed length $n=120$, resulting in a search space $\XXX$ of size $2^{120}$. The reward is specified by a set of modes $M \subset \XXX=\{0,1\}^n$ that is unknown to the learning agent. The reward of a generated sequence $x$ is defined in terms of Hamming distance $d$ from the modes: $R(x) = \exp\left(1 - n^{-1}\min_{y \in M} d(x,y)\right)$.  The vocabulary for the GFlowNets is $\{0,1\}$.  Most experiment settings are taken from \citet{malkin2022trajectory} and \citet{madan2022learning}.

Models are evaluated by tracking the number of modes according to the procedure in \citet{malkin2022trajectory} wherein we count a mode $m$ as ``discovered'' if we sample some $x$ such that $d(x,m) \leq \delta$.  The results are presented in Figure \ref{fig:bit_seq_1_modes_found} (mean and standard error are plotted over five random seeds).  We find that models trained with TS-GFN find ~60\% more modes than on-policy, tempering, and $\epsilon$-noisy.  TS-GFN soundly outperforms GAFN, whose pseudocount based exploration incentive is misaligned with the task's reward structure and seems to perform exploration in unhelpful regions of the (very large) search space.

\section{Conclusion}
\label{conclusion}
We have shown in this paper that using a Thompson sampling based exploration strategy for GFlowNets is a simple, computationally efficient, and performant alternative to prior GFlowNet exploration strategies.  We demonstrated how to adapt uncertainty estimation methods used for Thompson sampling in deep reinforcement learning to the GFlowNet domain and proved their efficacy on a grid and long sequence  generation task.  Finally, we believe that future work should involve trying TS-GFN on a wider array of experimental settings and building a theoretical framework for investigating sample complexity of GFlowNets.

\section*{Acknowledgments}

The authors acknowledge financial support from CIFAR, Genentech, IBM, Samsung, Microsoft, and Google.

\bibliography{gfn_ts}
\bibliographystyle{icml2023}

\newpage
\appendix
\onecolumn
\section{Additional algorithm details}
\label{app:algo}

\begin{algorithm}
\begin{algorithmic}

\STATE \textbf{Input}:
\STATE $\{P_{F,k}\}_{k=1}^K$: Family of $K$ different forward policies\;
\STATE $P_B$: Shared backwards policy\;
\STATE $\delta \in [0, 1]$: Parameter for the bootstrapping distribution (we assume the distribution is Bernoulli)\;
\STATE $L$: Loss function taking as input a trajectory $\tau$, forward policy $P_F$, and backward policy $P_B$
\FOR{each episode}
    \STATE Initialize $s_0$ from environment
    \STATE Initialize trajectory $\tau$ to empty list
    \STATE Pick forward policy to act with $P_{F,k}$ using $k \sim \textrm{Uniform}\{1,\dots,K\}$
    \FOR{step $t = 1, \dots$ until end of episode}
        \STATE Pick action $a_t \sim P_{F,k}(a_t|s_{t-1})$
        \STATE Receive state $s_t$ from environment
        \STATE Append $(s_t,a_t)$ to $\tau$
    \ENDFOR
    \STATE Add reward for trajectory $R(x)$ to $\tau$ using last state in $\tau$
    \STATE Sample bootstrap mask $m_k \sim \textrm{Bernoulli}(\delta)\quad \forall k$
    \STATE Compute loss $\ell = \sum_{k=1}^K m_k \cdot L(\tau, P_{F,k}, P_B)$
    \STATE Take gradient step on loss $\ell$
\ENDFOR
\end{algorithmic}

\caption{TS-GFN}
\label{algo:ts-gfn}
\end{algorithm}

\section{Experiment details: Grid}
\label{app:grid}
For brevity, we recall the definition of the reward function from Section \ref{exp:grid} as

\begin{equation*}
    R(x) = \sum_{k=1}^n \cos(2a_{k,1}\pi g(x_1)) + \sin(2 a_{k,2} \pi g(x_1)) + \cos(2b_{k,1}\pi g(x_2)) + \sin(2b_{k,2}\pi g(x_2)) 
\end{equation*}

The reward function was computed using the following hyperparameters.  The weights were set as $a_{k,1} = a_{k,2} = b_{k,1} = b_{k,2} = \frac{4k}{1000}$ with $n = 1000$ (the equivalent of \texttt{np.linspace(0, 4, 1000)}).  The grid side boundary constants were $c = -0.5, d= 0.5$ and the side length of the overall environment was $H = 64$ (so that the overall state space was of size $H \times H = 64^2 = 4096$).  Finally, we raised the reward by the exponent $\beta = 1.5$ so that we trained the GFlowNets using reward $R'(x) = R(x)^\beta$.

Besides the reward, architecture details are identical to those in \citet{malkin2022trajectory}, \citet{madan2022learning}, and \citet{bengio2021flow}.  The architecture of the forward and backward policy models are MLPs of the same architecture as in \citet{bengio2021flow}, taking a one-hot representation of the coordinates of $s$ as input and sharing all layers except the last.  The only difference comes from the TS-GFN implementation which has $K \cdot d$ heads for the output of the last layer where $d$ is the number of heads in the architecture of the non-TS-GFNs.

All models are trained with the Adam optimizer, the trajectory balance loss, and a batch size of 64 for a total of $400,000$ trajectories. Hyperparameters were tuned using the Optuna Bayesian optimization framework from project Ray \citep{akiba2019optuna,moritz2018ray}. Each method was allowed 100 hyperparameter samples from the Bayesian optimization procedure. We reported performance from the best hyperparameter setting found by the Bayesian optimization procedure averaged over five random seeds ($0,1,2,3,4$). We now detail the hyperparameters selected from the Bayesian optimization procedure for each exploration strategy.

For on-policy we found optimal hyperparameters of $0.00156$ for the model learning rate and $0.00121$ for the $\log Z$ learning rate. For tempering we found optimal hyperparameters of $0.00236$ for the model learning rate, $0.0695$ for the $\log Z$ learning rate, and $1.0458$ for the sampling policy temperature. For $\epsilon$-noisy we found optimal hyperparameters of $0.00112$ for the model learning rate, $0.0634$ for the $\log Z$ learning rate, and $0.00534$ for $\epsilon$.  For GAFN we found optimal hyperparameters of $0.000166$ for the model learning rate, $0.0955$ for the $\log Z$ learning rate, $0.144$ for the intrinsic reward weight, and the architecture of the RND networks were a 2 layer MLP with hidden layer dimension of $53$ and output embedding dimension of $96$ (the hidden layer dimension and embedding dimension were also tuned by the Bayesian optimization procedure). Finally, for Thompson sampling we found a model learning rate of $0.00266$, $\log Z$ learning rate of $0.0976$, ensemble size of $100$, bootstrap probability $\delta$ of $0.274$, and prior weight of $12.03$.
\section{Experiment details: Bit sequences}
\label{app:bit_seqs}
The modes $M$ as well as the test sequences are selected as described in \citet{malkin2022trajectory}. The policy for all methods is parameterized by a Transformer~\citep{vaswani17attn} with 3 layers, dimension 64, and 8 attention heads. All methods are trained for 50,000 iterations with minibatch size of 16 using Adam optimizer and the trajectory balance loss. 

All hyperparameters were tuned according to a grid search over the parameter values specified below. For on-policy we used a model learning rate of $0.0001$ picked from the set $\{0.0001, 0.001, 0.01\}$ and a $\log Z$ learning rate of $0.001$ from the set $\{0.001, 0.01\}$. For tempering we used a model learning rate of $0.0001$ picked from the set $\{0.0001, 0.001, 0.01\}$, a $\log Z$ learning rate of $0.001$ from the set $\{0.001, 0.01\}$, and sampling distribution temperature of $1.1$ from the set $\{1.05, 1.1, 1.25, 1.5\}$.  For $\epsilon$-noisy we used a learning rate of $0.001$ picked from the set $\{0.0001, 0.001, 0.01\}$, a $\log Z$ learning rate of $0.001$ from the set $\{0.001, 0.01\}$, and $\epsilon$ of $0.005$ from the set $\{0.01, 0.005, 0.001, 0.0005\}$.  For GAFN we used a learning rate of $0.001$ picked from the set $\{0.0001, 0.0005, 0.001\}$, a $\log Z$ learning rate of $0.1$ from the set $\{0.001, 0.01, 0.1\}$, an intrinsic reward weight of $0.5$ from the set $\{0.1, 0.5, 1.0, 5.0, 10.0\}$, the RND network was a 4 layer MLP with hidden layer dimension of $64$ and output dimension of $64$. For TS-GFN we used a model learning rate of $0.001$ picked from the set $\{0.0001, 0.001, 0.01\}$, a $\log Z$ learning rate of $0.001$ from the set $\{0.001, 0.01\}$, an ensemble size of $50$ picked from the set $\{10, 50, 100\}$, a prior weight of $4.0$ picked from the set $\{0.1, 1.0, 4.0\}$, and a bootstrap probability $\delta$ 0f $0.75$.

\end{document}